\documentclass[twoside,11pt]{article}

\usepackage{jmlr2e}

\firstpageno{1}

\begin{document}

\title{Scalable Bayesian Deep Learning with Kernel Seed Networks}

\author{\name Sam Maksoud \email s.maksoud@uqconnect.edu.au \\
       \addr School of Information Technology and Electrical Engineering\\
       The University of Queensland\\
       Brisbane, QLD 4072, AUS
       \AND
	   \name Kun Zhao \email k.zhao1@uq.edu.au \\
       \addr School of Information Technology and Electrical Engineering\\
       The University of Queensland\\
       Brisbane, QLD 4072, AUS
       \AND
	   \name Can Peng \email can.peng@uqconnect.edu.au \\
       \addr School of Information Technology and Electrical Engineering\\
       The University of Queensland\\
       Brisbane, QLD 4072, AUS
       \AND
	   \name Brian C. Lovell \email lovell@itee.uq.edu.au \\
       \addr School of Information Technology and Electrical Engineering\\
       The University of Queensland\\
       Brisbane, QLD 4072, AUS}
\editor{}

\maketitle

\begin{abstract}
This paper addresses the scalability problem of Bayesian deep neural networks. The performance of deep neural networks is undermined by the fact that these algorithms have poorly calibrated measures of uncertainty. This restricts their application in high risk domains such as computer aided diagnosis and autonomous vehicle navigation.  Bayesian Deep Learning (BDL) offers a promising method for representing uncertainty in neural network. However, BDL requires a separate set of parameters to store the mean and standard deviation of model weights to learn a distribution.  This results in a prohibitive 2-fold increase in the number of model parameters. To address this problem we present a method for performing BDL, namely Kernel Seed Networks (KSN), which does not require a 2-fold increase in the number of parameters. KSNs use 1x1 Convolution operations to learn a compressed latent space representation of the parameter distribution. In this paper we show how this allows KSNs to outperform conventional BDL methods while reducing the number of required parameters by up to a factor of 6.6.
\end{abstract}

\begin{keywords}
  Bayesian Networks, Weight Uncertainty, Optimization and learning methods, Deep Learning, Neural Networks
\end{keywords}

\section{Introduction}
\label{sect:Introduction}
Modern deep neural networks (DNNs) are capable of learning complex representations of high dimensional data \citep{resnet, vgg}. This has enabled DNNs to exceed human performance in a growing number of decision making tasks \citep{surpass_imagenet,surpass_atari,surpass_go}.
However, these algorithms are notoriously difficult to implement in real world decision making systems because conventional DNNs are prone to overfitting and thus have poorly calibrated measures of uncertainty~\citep{bbb}. A well-calibrated measure of uncertainty provides insight into the reliability of model predictions~\citep{calibration}. This is crucial for applications such as autonomous vehicle navigation~\citep{uncertain_bayes}, and computer assisted diagnosis~\citep{calibrate_medical}; where confidently incorrect decisions can potentially have fatal consequences. Hence, there has been a resurgent interest in exploring Bayesian methods for deep learning as they naturally provide well-calibrated representations of uncertainty.

Unlike conventional DNNs, which learn a fixed point estimate of model weights, Bayesian Neural Networks (BNNs) learn a weight distribution, parameterizing by the mean and standard deviation of the weights at each layer. The parameter distributions are optimized using variational sampling during training, which forces BNNs to be robust to perturbations of model weights.

Effectively, this makes BNN optimization equivalent to learning an infinite ensemble of models \citep{bbb}. With an ensemble, the variance of predictions among constituent models can be used to approximate the epistemic uncertainty \citep{ensemble}. Similarly, epistemic uncertainty in BNNs can be estimated using the variance of predictions among weights that are randomly sampled from the learned distributions.

While BNNs have proven to be effective at inducing uncertainty into the weights of a model, they require significantly more parameters than their DNN counterparts \citep{guide}. This makes it impractical to scale variational Bayesian learning to the depth of modern DNN architectures. To overcome this problem, \cite{mcdropout} proposed the use of Monte-Carlo Dropout (MC-Drop); a method for approximating Bayesian inference by applying Dropout \citep{dropout} during both model training and inference.

A major advantage of MC-Drop, is that it requires no additional parameters to approximate Bayesian inference. MC-Drop represents model uncertainty by sampling sub-networks with randomly dropped out units and connections to perform variational inference \citep{mcdropout}. Although this approach captures the uncertainty of model activations, it is unable to represent the uncertainty of model weights since all sub-network perturbations sampled using the Dropout method will share the same parameters \citep{dropout}. This limits the scope in which MC-Drop can be applied, since many applications such as continual learning \citep{continual} and model pruning \citep{prune, bbb}, require representations of uncertainty over the entire parameter space. To address these limitations, \cite{vogn} propose the Variational Online Gauss-Newton (VOGN) algorithm. VOGN parameterizes a distribution of weights by modelling the output of a DNN as the mean, and the $2^{nd}$ raw moment uncentered variance of computed gradients as the variance.

Similar to BNNs, VOGN optimization involves randomly sampling a set of parameters from a weight distribution to compute variational gradients, and perform variational inference. However, the key advantage of VOGN optimization, is that it does not require learning a separate set of mean and standard deviation parameters, since it models the variance with a vector that is already used in adaptive optimization methods such as Adam \citep{adam}. But it is important to note that VOGN optimization only finds an approximate solution to the BNN objective \citep{bbb}. \cite{EWCvVCL} demonstrate how this approximation can result in a trade-off between accuracy and ease of implementation when applied to Variational Continual Learning (VCL) methods \citep{vcl}. It was shown that using the VOGN approximation for VCL, results in worse performance in certain tasks comparing to conventional VCL, which optimizes the variance of the weight distributions directly \citep{vcl}, rather than employing the online estimate of the diagonal of the Fisher Information Matrix to update the variance of the parameters \citep{EWCvVCL,vogn}. Hence, despite the high computational cost of BNNs, their are benefits to directly optimizing the mean and standard deviation parameters of a weight distribution during training.

In this paper, we introduce a novel class of BNNs, namely, Kernel Seed Networks (KSNs), which significantly reduce the number of parameters required for Bayesian deep learning (BDL). In contrast with conventional BNNs, which require doubling the size of the network in order to learn a separate set of mean and standard deviations weights for each of the parameter distributions, our method applies $1\times1$ convolutional kernels to trainable \textit{``seed''} vectors to decode the mean and standard deviation weight vectors. The ability of $1\times1$ convolutional layers to map inputs to a higher-dimensional space allows for further reduction of parameters, since they can be used to upsample downsized seed kernels to the required filter size. In our experiments on MNIST~\citep{mnist}, FMNIST~\citep{fmnist} and CIFAR~\citep{cifar} datasets, we demonstrate how our proposed method can be used to conduct BDL with a $44$--$70\%$ reduction in the number of parameters (compared to conventional BNNs) without compromising on model performance.

\label{methods}
\begin{figure}
\begin{center}
   \includegraphics[width=0.9\linewidth]{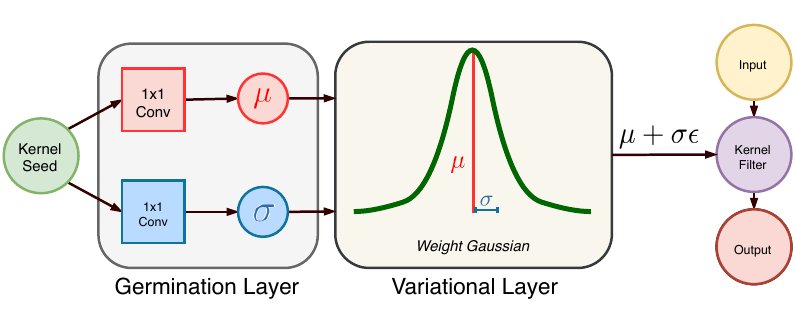}
\end{center}
   \caption{Framework of the proposed self-germinating kernel filter. The $N$ dimensional kernel seed vector is passed through the germination layer to decode the mean $\mu$ and standard deviations $\sigma$ for the layer weight distribution. In the variational layer, $\mu$ and $\sigma$ are used to respectively shift and scale a randomly sampled unit Gaussian vector $\epsilon$. This process returns the final $N$ dimensional kernel filter (purple) that is applied to the inputs of the layer.}
\label{fig:model}
\end{figure}

\section{Kernel Seed Networks}

As described in Section \ref{sect:Introduction}, the purpose of KSNs is to reduce the number of parameters required to construct the BNNs first described by \cite{bbb}. To this end, we propose to replace the $N$ dimensional conventional kernel filters in DNNs with self-germinating kernel filters (Figure \ref{fig:model}) comprising three main components: (1) an $N$ dimensional kernel seed, (2) a kernel germination layer, and (3) a variational kernel layer. We describe the details of these components and our optimization protocol below.

\subsection{The Kernel Seed}
\label{seed layer}

The purpose of the kernel seed is to store a latent-space representation $\psi$ of the layer weight distribution $W$. To this end, we construct seed kernels for all linear and convolutional layers in a neural network as follows.

\noindent
\textbf{Linear Kernel Seeds.} A Linear kernel seed $\psi_{FC}$ stores the compressed weight distributions for a given linear transformation layer $K_{FC} \in \mathbb{R}^{C_{in} \times C_{out}}$, where $C_{in}$ and $C_{out}$ are the number of input channels and output channels respectively. Since the kernel germination method is capable of mapping kernel seeds to a higher dimensional space, we can reduce the dimensions of $\psi_{FC}$ by applying the scaling parameter $\delta$ to $C_{f}=\min(C_{in}, C_{out})$. Specifically, the linear kernel seed can be expressed as $\psi_{FC} \in \mathbb{R}^{C_{pip} \times C_{F}}$, where $C_{pip} = \delta C_{f}$ and $C_{F}=\max(C_{in}, C_{out})$.

\noindent
\textbf{Convolutional Kernel Seeds.} A convolutional kernel seed $\psi_{C}$ stores the compressed weight distributions for a given convolutional layer $K_{C} \in \mathbb{R}^{C_{in} \times C_{out} \times k \times k}$, where $k \times k$ are the dimensions of the convolving kernel filter. As with linear kernel seeds, we can reduce the dimensions of $\psi_{C}$ by applying the scaling parameter $\delta$ to $C_{f}$ yielding $C_{pip}$. Specifically, the convolutional kernel seed can be expressed as $\psi_{C} \in \mathbb{R}^{C_{pip} \times C_{F} \times k \times k}$.

\begin{figure}
\begin{center}
   \includegraphics[width=\linewidth]{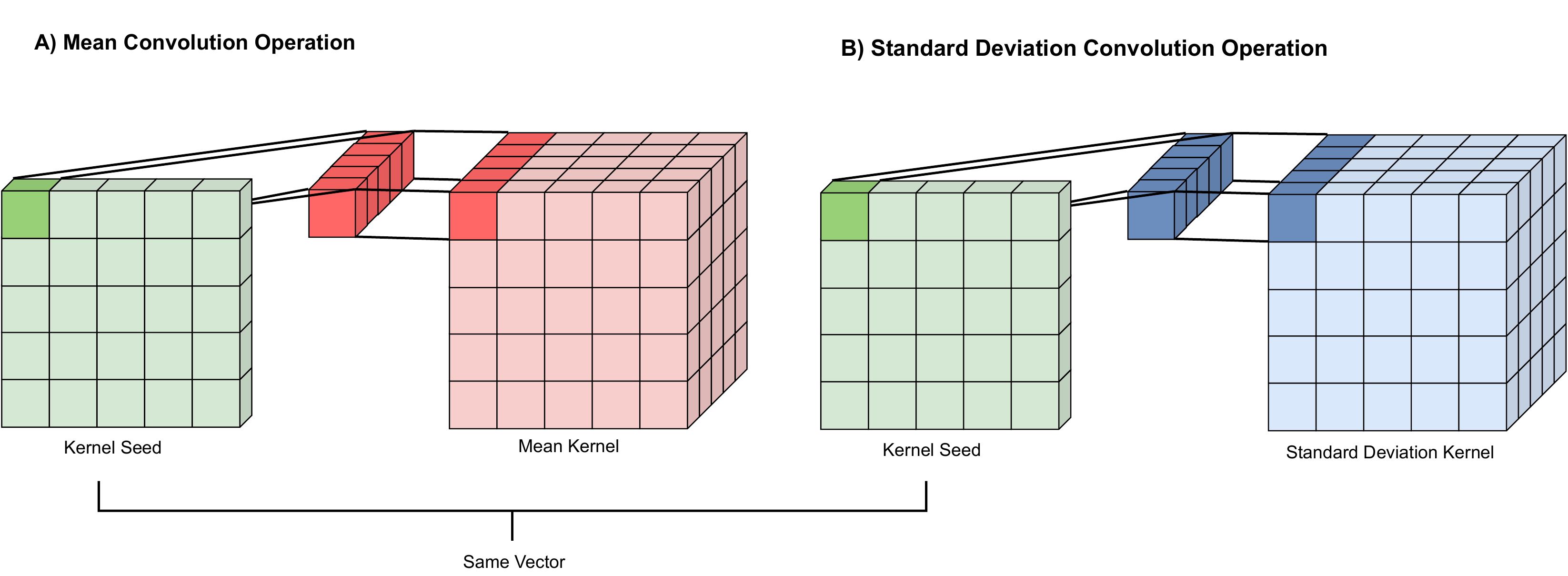}
\end{center}
   \caption{The image above illustrates the germination procedure. Specifically, two $1$x$1$ convolutional upsampling operations are applied to a compressed seed vector to decode the mean and standard deviation parameters of the weight distribution. In effect, the kernel seed is a latent space representation of the weight distribution from which the parameters of a given layer are derived.}
\label{fig:germ}
\end{figure}

\subsection{Germination Layer}
\label{germ layer}
The germination layer comprises two convolutional decoders (germinators): (1) a $\mu$ decoder, and (2) a $\rho$ decoder. It receives the kernel seed $\psi$ as input and decodes the $\mu$ and $\rho$ parameters (Figure~\ref{fig:germ}), which are then used to reconstruct $W$.

\noindent
\textbf{Linear Seed Germinators.} To decode the $\mu$ and $\rho$ parameters for a linear transformation layer $K_{FC}$, we apply two convolutional germinating kernels, $G_{FC_{\mu}}$ and $G_{FC_{\rho}}$, to the linear kernel seed $\psi_{FC}$. $G_{FC_{\mu}}$ and $G_{FC_{\rho}}$ are essentially $1$D convolutional kernels where $G_{FC} \in \mathbb{R}^{C_{f} \times C_{pip} \times 1}$. When applied to $\psi_{FC}$, $G_{FC}$ outputs a matrix of size $C_{F} \times C_{f}$. When $C_{F}=C_{out}$, a transformation operation is applied to the outputs of $G_{FC_{\mu}}$ and $G_{FC_{\rho}}$ yielding $W_{\mu} \in \mathbb{R}^{C_{in} \times C_{out}}$ and $W_{\rho} \in \mathbb{R}^{C_{in} \times C_{out}}$ respectively.

\noindent
\textbf{Convolutional Seed Germinators.} To decode the $\mu$ and $\rho$ parameters for a convolutional layer $K_{C}$, we apply two convolutional germinating kernels, $G_{C_{\mu}}$ and $G_{C_{\rho}}$, to the convolutional kernel seed $\psi_{C}$. $G_{C_{\mu}}$ and $G_{C_{\rho}}$ are essentially $2$D convolutional kernels where $G_{C} \in \mathbb{R}^{C_{f} \times C_{pip} \times 1 \times 1}$. When applied to $\psi_{C}$, $G_{C}$ outputs a tensor of size $C_{F} \times C_{f} \times k \times k$. As with linear kernel seeds, when $C_{F}=C_{out}$, a transformation operation is applied to the outputs of $G_{C_{\mu}}$ and $G_{C_{\rho}}$ yielding $W_{\mu} \in \mathbb{R}^{C_{in} \times C_{out} \times k \times k}$ and $W_{\rho} \in \mathbb{R}^{C_{in} \times C_{out} \times k \times k}$ respectively.

\subsection{Variational Kernel Layer}
\label{var layer}
The purpose of the variational kernel layer is to randomly sample a subset of weights $w \in W$. As described by \cite{bbb}, the subset of weights can be obtained by using the mean $\mu$ and standard deviation $\sigma = \log(1 + \exp(\rho))$ parameters to respectively shift and scale a randomly sampled noise vector $\epsilon \sim \mathcal{N}(0,1)$ yielding $w = \mu + \sigma\epsilon$. The subset of weights $w$ is then applied to the inputs of the layer as illustrated in Figure \ref{fig:model}.

\subsection{Optimization Protocol}
\label{scale mix}

While KSNs differ from BNNs in the way $\mu$ and $\sigma$ parameters are stored, the process of optimizing the distributions of model weights is the same. Thus, for KSN model training, we follow the BNN optimization protocol described by \cite{bbb}; using a scale mixture Gaussian prior:

\begin{equation}
\label{eq:output}
P(w)=\prod_{i} \pi \mathcal{N}\left(w_{i} \mid 0, \sigma_{1}^{2}\right)+(1-\pi) \mathcal{N}\left(w_{i} \mid 0, \sigma_{2}^{2}\right)
\end{equation}

\noindent where $w_{i}$ is the $i^{th}$ weight, $\pi = \frac{1}{4}$, $-\log \sigma_{1}^{2} = 0$ and $-\log \sigma_{2}^{2} = 6$ are the variances of the mixture components in the Gaussian prior. We use a fixed learning rate of $1e^{-3}$ for our experiments and initialize the weights of all seed kernels using Glorot uniform initialization \citep{glorot}.

\begin{table}
\small
\begin{center}
\begin{tabular}{|c|c|c|c|c|c|c|c|c|c|}
\hline

\textbf{Model} & \textbf{$\delta$} & \textbf{DR} & \textbf{Params} & \textbf{RS} & \textbf{ACC} $\uparrow$  & \textbf{NLL} $\downarrow$ & \textbf{ECE} $\downarrow$ & \textbf{ACE} $\downarrow$ & \textbf{MCE} $\downarrow$ \\ \hline
Baseline & - & - & $11.2$M & $1.00$ & $0.9894$ & $0.0353$ & $0.0021$ & $0.0767$ & $0.2432$ \\
Baseline & - & $0.10$ & $11.2$M & $1.00$ & $0.9910$ & $0.0268$ & $0.0018$ & $0.1392$ & $0.6413$ \\  \hline
MC-Drop  & - & $0.10$ &  $11.2$M & $1.00$ & $0.9910$ & $0.0269$ & $0.0021$ & $0.0688$ & $0.2845$ \\  \hline
VOGN & - & - &  $11.2$M & $1.00$ & $0.9921$ & $1.4763$ & $0.7626$ & $0.6688$ & $0.7673$ \\  \hline
KSN-E & $1.00$ & - & $13.6$M & $1.22$ & $0.9903$ & $0.0417$ & $0.0174$ & $0.1306$ & $0.2004$ \\
KSN-P & $1.00$ & - & $13.6$M & $1.22$ & $\textbf{0.9926}$ & $\textbf{0.0203}$ & $0.0018$ & $0.1199$ & $0.3479$ \\
FKSN & $1.00$ & - & $12.4$M & $1.11$ & $0.9925$ & $0.0250$ & $\textbf{0.0016}$ & $0.1014$ & $0.3775$ \\  \hline
KSN-E & $0.75$ & - & $10.2$M & $0.91$ & $0.9904$ & $0.0373$ & $0.0108$ & $0.1342$ & $0.2639$ \\
KSN-P & $0.75$ & - & $10.2$M & $0.91$ & $0.9909$ & $0.0313$ & $0.0030$ & $0.1894$ & $0.7315$ \\
FKSN & $0.75$ & - & $9.3$M & $0.83$ & $0.9887$ & $0.0355$ & $0.0018$ & $0.0604$ & $0.2253$ \\  \hline
KSN-E & $0.50$ & - & $6.8$M & $0.61$ & $0.9858$ & $0.0591$ & $0.0199$ & $0.1377$ & $0.2750$ \\
KSN-P & $0.50$ & - & $6.8$M & $0.61$ & $0.9851$ & $0.0506$ & $0.0039$ & $0.1153$ & $0.3541$ \\
FKSN & $0.50$ & - & $6.2$M & $0.56$ & $0.9692$ & $0.1042$ & $0.0124$ & $0.0866$ & $0.2254$ \\  \hline
KSN-E & $0.25$ & - & $3.4$M & $0.31$ & $0.9845$ & $0.0779$ & $0.0286$ & $0.1401$ & $0.3097$ \\
KSN-P & $0.25$ & - & $3.4$M & $0.31$ & $0.9873$ & $0.0654$ & $0.0082$ & $0.1083$ & $0.1871$ \\
FKSN & $0.25$ & - & $3.1$M & $0.28$ & $0.9679$ & $0.1828$ & $0.0054$ & $\textbf{0.0407}$ & $\textbf{0.0823}$ \\  \hline
BNN-E & - & - & $22.3$M & $2.00$ & $0.9732$ & $0.2132$ & $0.1299$ & $0.2021$ & $0.3227$ \\
BNN-P & - & - & $22.3$M & $2.00$ & $0.9898$ & $0.0359$ & $0.0032$ & $0.1167$ & $0.3767$ \\   \hline

\end{tabular}
\end{center}
\caption{Model performance on the MNIST dataset \citep{mnist}.}
\label{table:MNIST}
\end{table}

\begin{table}
\small
\begin{center}
\begin{tabular}{|c|c|c|c|c|c|c|c|c|c|}
\hline

\textbf{Model} & \textbf{$\delta$}  & \textbf{DR} & \textbf{Params} & \textbf{RS} &  \textbf{ACC} $\uparrow$  & \textbf{NLL} $\downarrow$ & \textbf{ECE} $\downarrow$ & \textbf{ACE} $\downarrow$ & \textbf{MCE} $\downarrow$ \\ \hline
Baseline & - & - & $11.2$M & $1.00$ & $0.9046$ & $0.2641$ & $0.0222$ & $0.0792$ & $0.2981$ \\
Baseline & - & $0.10$ & $11.2$M & $1.00$ & $0.9082$ & $0.2601$ & $0.0198$ & $0.0655$ & $0.2762$ \\  \hline
MC-Drop & - & $0.10$ & $11.2$M & $1.00$ & $0.9067$ & $0.2570$ & $0.0140$ & $0.0373$ & $0.1129$ \\  \hline
VOGN & - & - & $11.2$M & $1.00$ & $0.8688$ & $1.6317$ & $0.6596$ & $0.5971$ & $0.7439$ \\  \hline
KSN-E & $1.00$ & - & $13.6$M & $1.22$ & $0.9048$ & $0.2708$ & $0.0232$ & $\textbf{0.0338}$ & $\textbf{0.0708}$ \\
KSN-P & $1.00$ & - & $13.6$M & $1.22$ & $0.9058$ & $0.2695$ & $0.0209$ & $0.0656$ & $0.2144$ \\
FKSN & $1.00$ & - &  $12.4$M & $1.11$ & $0.9122$ & $0.2463$ & $0.0206$ & $0.0733$ & $0.2804$ \\  \hline
KSN-E & $0.75$ & - & $10.2$M & $0.91$ & $0.9094$ & $0.2583$ & $0.0190$ & $0.0536$ & $0.1610$ \\
KSN-P & $0.75$ & - & $10.2$M & $0.91$ & $\textbf{0.9175}$ & $\textbf{0.2388}$ & $0.0205$ & $0.0847$ & $0.2768$ \\
FKSN & $0.75$ & - & $9.3$M & $0.83$ & $0.8938$ & $0.2864$ & $\textbf{0.0078}$ & $0.0378$ & $0.1433$ \\  \hline
KSN-E & $0.50$ & - & $6.8$M & $0.61$ & $0.8785$ & $0.3560$ & $0.0248$ & $0.1332$ & $0.8095$ \\
KSN-P & $0.50$ & - & $6.8$M & $0.61$ & $0.8893$ & $0.3115$ & $0.0256$ & $0.0535$ & $0.1354$ \\
FKSN & $0.50$ & - & $6.2$M & $0.56$ & $0.8944$ & $0.2851$ & $0.0174$ & $0.0448$ & $0.1234$ \\  \hline
KSN-E & $0.25$ & - & $3.4$M & $0.31$ & $0.8724$ & $0.3910$ & $0.0530$ & $0.0908$ & $0.1846$ \\
KSN-P & $0.25$ & - & $3.4$M & $0.31$ & $0.8777$ & $0.3967$ & $0.0218$ & $0.0557$ & $0.1842$ \\
FKSN & $0.25$ & - & $3.1$M & $0.28$ & $0.8744$ & $0.4068$ & $0.0134$ & $0.0586$ & $0.1982$ \\  \hline
BNN-E & - & - & $22.3$M & $2.00$ & $0.8374$ & $0.5023$ & $0.0659$ & $0.0920$ & $0.1897$ \\
BNN-P & - & - & $22.3$M & $2.00$ & $0.8934$ & $0.3149$ & $0.0361$ & $0.0744$ & $0.1167$ \\   \hline

\end{tabular}
\end{center}
\caption{Model performance on the FMNIST dataset \citep{fmnist}.}
\label{table:FMNIST}
\end{table}

\begin{table}
\small
\begin{center}
\begin{tabular}{|c|c|c|c|c|c|c|c|c|c|}
\hline

\textbf{Model} & \textbf{$\delta$}  & \textbf{DR} & \textbf{Params} & \textbf{RS} & \textbf{ACC} $\uparrow$  & \textbf{NLL} $\downarrow$ & \textbf{ECE} $\downarrow$ & \textbf{ACE} $\downarrow$ & \textbf{MCE} $\downarrow$ \\ \hline
Baseline & - & - & $11.2$M & $1.00$ & $0.9030$ & $0.6497$ & $0.0768$ & $0.1884$ & $0.3399$ \\
Baseline & - & $0.10$ & $11.2$M & $1.00$ & $\textbf{0.9188}$ & $0.4685$ & $0.0577$ & $0.1720$ & $0.2903$ \\  \hline
MC-Drop & - & $0.10$ & $11.2$M & $1.00$ & $0.9165$ & $\textbf{0.3458}$ & $0.0346$ & $0.0623$ & $0.1285$ \\  \hline
VOGN & - & - & $11.2$M & $1.00$ & $0.7615$ & $1.7079$ & $0.5463$ & $0.4468$ & $0.6208$ \\  \hline
KSN-E & $1.00$ & - & $13.6$M & $1.22$ & $0.6143$ & $1.0430$ & $\textbf{0.0178}$ & $\textbf{0.0208}$ & $\textbf{0.0590}$ \\
KSN-P & $1.00$ & - & $13.6$M & $1.22$ & $0.8531$ & $0.6044$ & $0.0913$ & $0.1235$ & $0.3179$ \\
FKSN & $1.00$ & - & $12.4$M & $1.11$ & $0.9058$ & $0.5861$ & $0.0694$ & $0.1843$ & $0.2875$ \\  \hline
KSN-E & $0.75$ & - & $10.2$M & $0.91$ & $0.7844$ & $0.7708$ & $0.1830$ & $0.1633$ & $0.2861$ \\
KSN-P & $0.75$ & - & $10.2$M & $0.91$ & $0.8821$ & $0.4120$ & $0.0507$ & $0.1127$ & $0.2077$ \\
FKSN & $0.75$ & - & $9.3$M & $0.83$ & $0.8912$ & $0.6348$ & $0.0792$ & $0.1876$ & $0.3479$ \\  \hline
KSN-E & $0.50$ & - & $6.8$M & $0.61$ & $0.7985$ & $0.7071$ & $0.1475$ & $0.1418$ & $0.2237$ \\
KSN-P & $0.50$ & - & $6.8$M & $0.61$ & $0.8860$ & $0.5057$ & $0.0661$ & $0.1300$ & $0.2123$ \\
FKSN & $0.50$ & - & $6.2$M & $0.56$ & $0.8689$ & $0.7621$ & $0.0941$ & $0.1965$ & $0.3579$ \\  \hline
KSN-E & $0.25$ & - & $3.4$M & $0.31$ & $0.8219$ & $0.5765$ & $0.0869$ & $0.1326$ & $0.4865$ \\
KSN-P & $0.25$ & - & $3.4$M & $0.31$ & $0.8382$ & $0.7951$ & $0.0637$ & $0.1355$ & $0.1973$ \\
FKSN & $0.25$ & - & $3.1$M & $0.28$ & $0.8476$ & $0.9671$ & $0.0879$ & $0.1830$ & $0.2998$ \\  \hline
BNN-E & - & - & $22.3$M & $2.00$ & $0.3905$ & $1.6020$ & $0.0935$ & $0.0841$ & $0.1840$ \\
BNN-P & - & - & $22.3$M & $2.00$ & $0.5596$ & $1.6792$ & $0.2432$ & $0.1928$ & $0.3575$ \\   \hline

\end{tabular}
\end{center}
\caption{Model performance on the CIFAR10 dataset \citep{cifar}.}
\label{table:CIFAR10}
\end{table}

\begin{table}
\small
\begin{center}
\begin{tabular}{|c|c|c|c|c|c|c|c|c|c|}
\hline

\textbf{Model} & \textbf{$\delta$}  & \textbf{DR} & \textbf{Params} & \textbf{RS} & \textbf{ACC} $\uparrow$  & \textbf{NLL} $\downarrow$ & \textbf{ECE} $\downarrow$ & \textbf{ACE} $\downarrow$ & \textbf{MCE} $\downarrow$ \\ \hline
Baseline & - & - & $11.2$M & $1.00$ & $0.6541$ & $2.9428$ & $0.2609$ & $0.3131$ & $0.4993$ \\
Baseline & - & $0.10$ & $11.2$M & $1.00$ & $\textbf{0.6700}$ & $2.4747$ & $0.2356$ & $0.2786$ & $0.4587$ \\  \hline
MC-Drop & - & $0.10$ & $11.2$M & $1.00$ & $0.6675$ & $1.8902$ & $0.1509$ & $0.1759$ & $0.2460$ \\  \hline
VOGN & - & - & $11.2$M & $1.00$ & $0.4665$ & $4.1917$ & $0.4458$ & $0.4458$ & $0.4458$ \\  \hline
KSN-E & $1.00$ & - & $13.7$M & $1.22$ & $0.4947$ & $2.0648$ & $0.1536$ & $0.1529$ & $0.2418$ \\
KSN-P & $1.00$ & - & $13.7$M & $1.22$ & $0.6235$ & $1.9624$ & $0.2115$ & $0.2202$ & $0.3861$ \\
FKSN & $1.00$ & - & $12.6$M & $1.11$ & $0.654$ & $2.6108$ & $0.2534$ & $0.3021$ & $0.4760$ \\  \hline
KSN-E & $0.75$ & - & $10.3$M & $0.91$ & $0.5659$ & $\textbf{1.7881}$ & $0.1632$ & $0.1497$ & $0.2697$ \\
KSN-P & $0.75$ & - & $10.3$M & $0.91$ & $0.6223$ & $1.9163$ & $0.2045$ & $0.2204$ & $0.3710$ \\
FKSN & $0.75$ & - & $9.3$M & $0.83$ & $0.6390$ & $2.6004$ & $0.2543$ & $0.2937$ & $0.4748$ \\  \hline
KSN-E & $0.50$ & - & $6.8$M & $0.61$ & $0.5130$ & $1.9858$ & $0.0864$ & $0.0926$ & $0.1713$ \\
KSN-P & $0.50$ & - & $6.8$M & $0.61$ & $0.5607$ & $2.0893$ & $0.1941$ & $0.1923$ & $0.3208$ \\
FKSN & $0.50$ & - & $6.2$M & $0.56$ & $0.5720$ & $3.2770$ & $0.3046$ & $0.3149$ & $0.5168$ \\  \hline
KSN-E & $0.25$ & - & $3.4$M & $0.31$ & $0.5075$ & $1.9962$ & $\textbf{0.0358}$ & $\textbf{0.0450}$ & $\textbf{0.0931}$ \\
KSN-P & $0.25$ & - & $3.4$M & $0.31$ & $0.5426$ & $2.2156$ & $0.2064$ & $0.1996$ & $0.3763$ \\
FKSN & $0.25$ & - & $3.1$M & $0.28$ & $0.5492$ & $3.4902$ & $0.3151$ & $0.3168$ & $0.5109$ \\  \hline
BNN-E & - & - & $22.4$M & $2.00$ & $0.2714$ & $3.0345$ & $0.0667$ & $0.1519$ & $0.3081$ \\
BNN-P & - & - & $22.5$M & $2.00$ & $0.4315$ & $2.2201$ & $0.1199$ & $0.1216$ & $0.1898$ \\   \hline

\end{tabular}
\end{center}
\caption{Model performance on the CIFAR100 dataset \citep{cifar}.}
\label{table:CIFAR100}
\end{table}

\begin{figure}[!htb]
	
   \begin{minipage}{0.49\textwidth}
     \centering
     \includegraphics[width=0.8\linewidth]{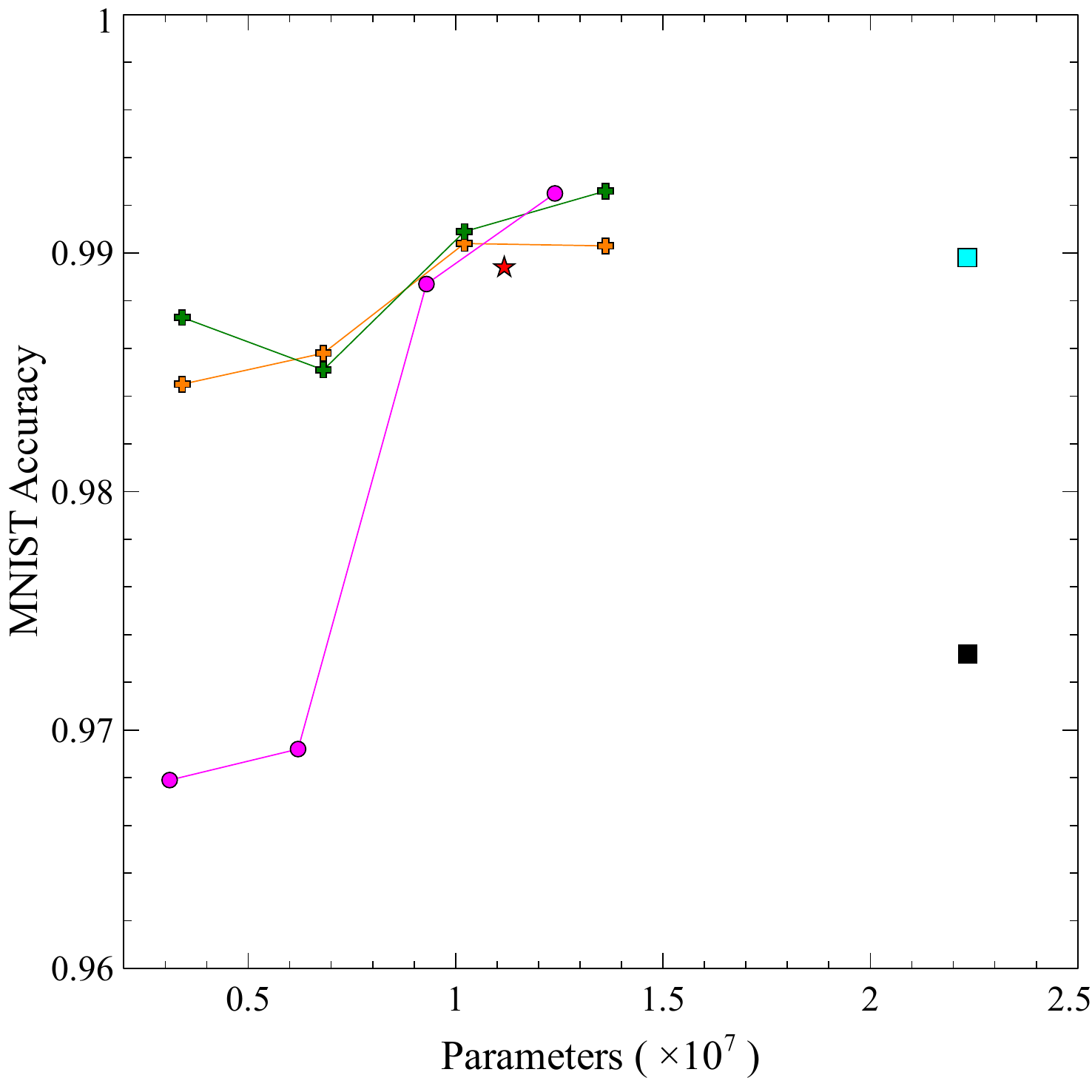}
   \end{minipage}\hfill
   \begin{minipage}{0.49\textwidth}
     \centering
     \includegraphics[width=0.8\linewidth]{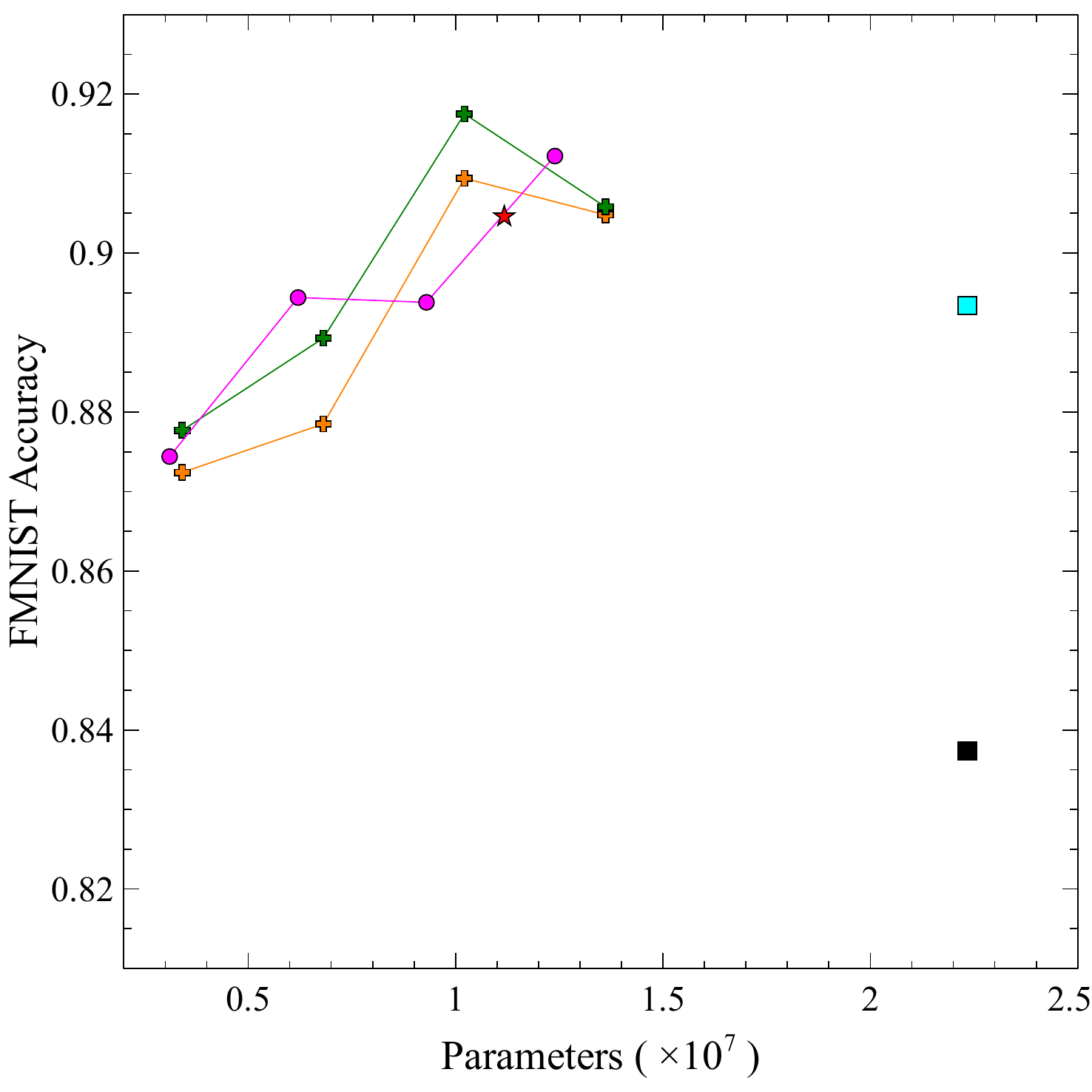}
   \end{minipage}

   \begin{minipage}{0.49\textwidth}
     \centering
     \includegraphics[width=0.8\linewidth]{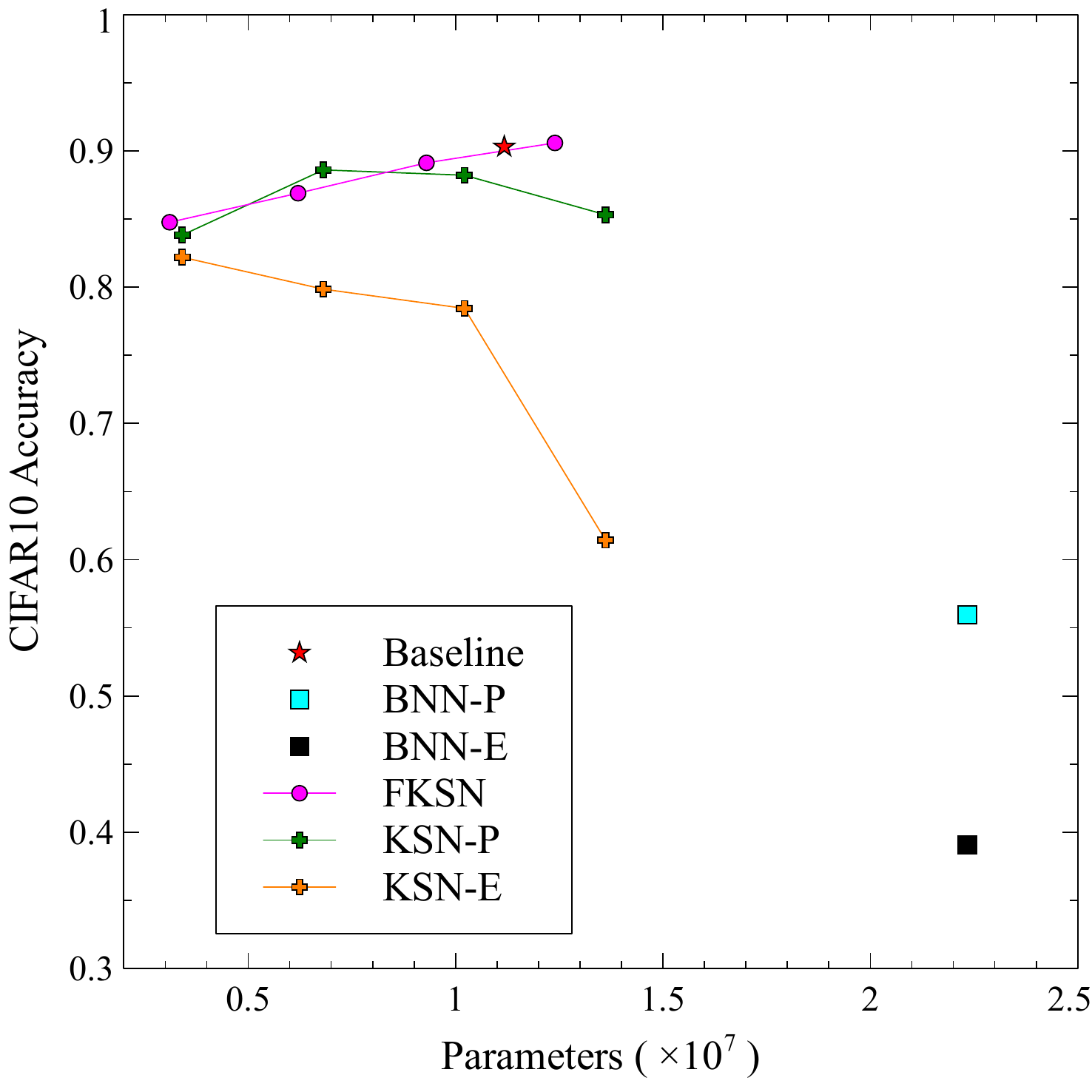}
   \end{minipage}\hfill
   \begin{minipage}{0.49\textwidth}
     \centering
     \includegraphics[width=0.8\linewidth]{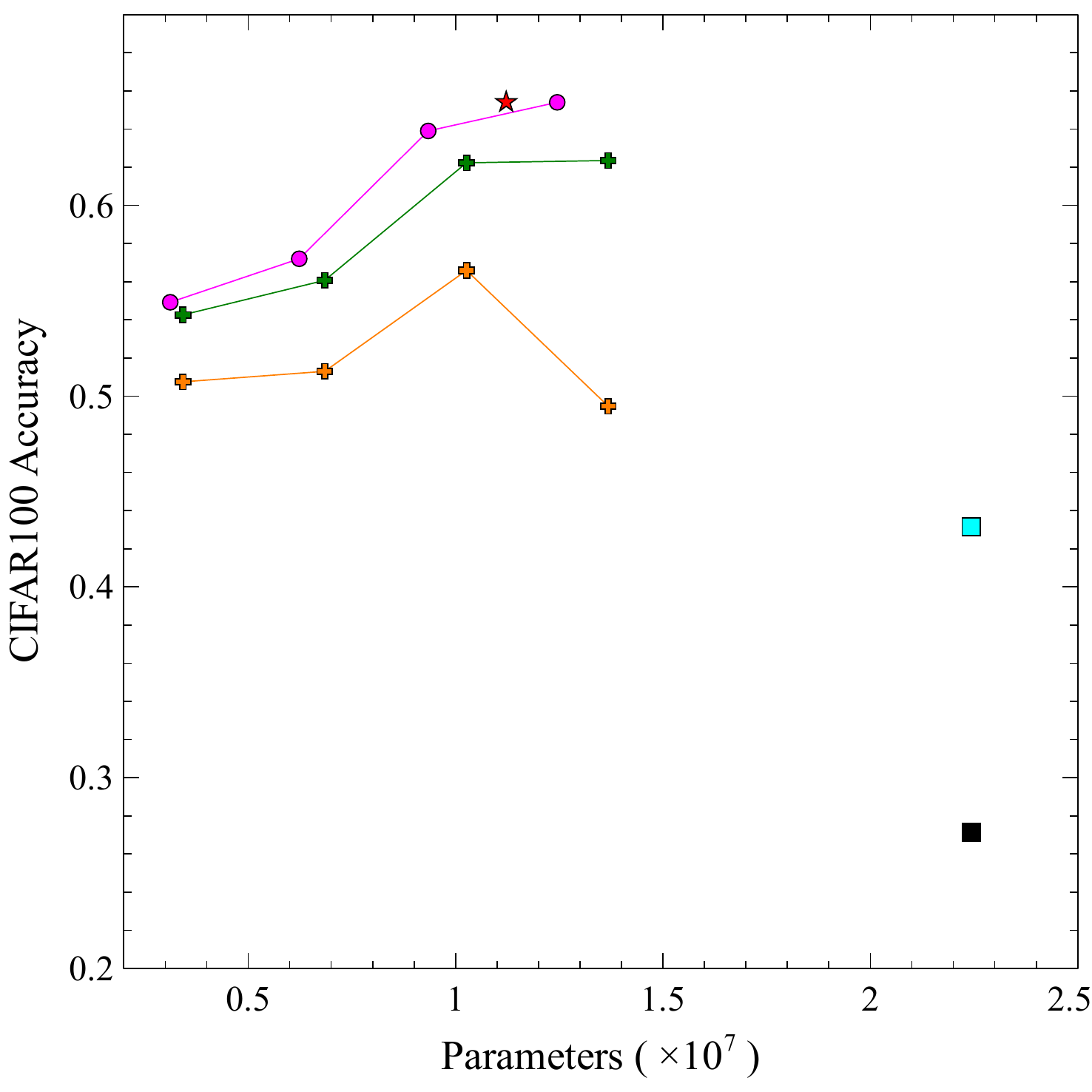}
   \end{minipage}

  \caption{Model accuracy vs. number of parameters on MNIST, FMNIST, CIFAR10 and CIFAR100 classification tasks. Compared with BNNs, KSN-E and KSN-P methods consistently achieve high classification accuracy using significantly fewer parameters.}
  \setcounter{figure}{1}
  \label{fig:results}
\end{figure}

\section{Experiments}
\label{sec:experiments}

The aim of the KSN is to directly optimize a set of $\mu$ and $\rho$ for a weight distribution without a 2-fold increase in the number of parameters, as required by conventional BNNs \citep{bbb, guide}. Thus, to validate the effectiveness of the KSN we evaluate the relationship between model size and classification accuracy. Specifically, in Figure \ref{fig:results} we observe how changes in the number of model parameters affects model performance on MNIST, FMNIST, CIFAR10 and CIFAR100 classification tasks. Furthermore, in Tables \ref{table:MNIST}, \ref{table:FMNIST}, \ref{table:CIFAR10}, and \ref{table:CIFAR100}, we evaluate how the number of parameters (Params) and model size relative to a ResNet18 baseline (RS) affects the Accuracy (ACC), Negative Log Loss (NLL), Expected Calibration Error (ECE), Average Calibration Error (ACE) and Maximum Calibration Error (MCE) rates when performing variational inference. A well calibrated model should have low NLL, ECE, ACE and MCE rates on unseen test data~\citep{bayesprincipals}. Hence, they provide a means of comparing each method's ability to produce well calibrated models with Bayesian inference.

\subsection{Method Comparison}

The details of each of the methods used in our experiments are described below. For the simple MNIST and FMNIST classification tasks, accuracy was evaluated after training for $5$ epochs. For the more complex CIFAR$10$ and CIFAR$100$ classification tasks, accuracy was evaluated after training for $100$ epochs.

\noindent\textbf{Baseline Fixed-Point Estimate.} All models in our experiments use a ResNet18 backbone \citep{resnet}. In order to observe the effects of inference using distributions of model weights, we require a baseline fixed point estimate for comparison. Thus, we establish a performance benchmark by modelling fixed point estimates of model weights for each classification task with ResNet18.

\noindent\textbf{Monte-Carlo Dropout.} We implement the MC-Drop method as introduced by \cite{mcdropout}. Following the work of \cite{bayesprincipals}, we configure MC-Drop with a dropout rate (DR) of $0.1$. In line with other methods used in our study, we use a ResNet18 backbone \citep{resnet} and also provide the posterior (Baseline with DR $= 0.1$), where we perform static inference on a model trained with dropout.

\noindent\textbf{Variational Online Gauss-Newton.} To compare BNN optimization with variational natural gradient optimization, we implement the VOGN optimization protocol introduced by \cite{vogn}. Specifically, we train a ResNet18 \citep{resnet} backbone with Vadam optimization \citep{vogn} and randomly sample $10$ weight perturbations to compute the variational gradients during training, and perform the variational inference.

\noindent\textbf{Bayesian Neural Networks.} The BNNs used in these experiments are constructed by essentially cloning the ResNet18 backbone such that one set of parameters stores the $\mu$, and the other stores the $\rho$ for the distribution of model weights. During training, the BNNs are optimized using \textit{bayes-by-backprop} (BBB) \citep{bbb}. The original implementation of BBB could only be applied to linear transformation layers, however, recent work by \cite{guide} provides a generalization of BBB that can be applied to convolutional layers.  During inference, we sample the posterior mean of the learned distribution (BNN-P), as well as an ensemble of $10$ randomly sampled network weights (BNN-E).

\noindent\textbf{Fixed-Point Kernel Seed Networks.} While the Baseline ResNet18 method is useful to study the effects of BDL on conventional DNNs, the KSN method requires its own fixed-point implementation for comparison. This is because the process of reconstructing a layer from a kernel seed matrix is unique to KSNs. Thus, when comparing the performance of KSNs to the Baseline ResNet18 method, we cannot distinguish whether differences in model performance are due to the effects of BDL on KSNs or the fundamental variations between DNN and KSN architectures.
Thus, we also provide the evaluation of the Fixed-Point Kernel Seed Network (FKSN). The Fixed-Point Kernel Seed Network is similar to the Variational Kernel Seed Network described in Section \ref{methods}. However, the F-KSN does not model distributions of weights, but rather, it simply reconstructs the $\mu$ weight for each layer. Specifically, $G_{C_{\mu}}$ germinates $\psi_{C}$ directly to $K_{C}$ for convolutional layers, and $G_{FC_{\mu}}$ germinates $\psi_{FC}$ directly to $K_{FC}$ for linear transformation layers, without variational sampling. This enables us to observe the effects of applying BDL to KSNs, as FKSNs are a fixed-point implementation of our KSN model.

\noindent\textbf{Variational Kernel Seed Networks.} We implement the Variational Kernel Seed Network as described in Section \ref{methods}, to germinate a neural network with a ResNet18 architecture \citep{resnet}. As with BNNs, we provide the posterior mean of the learned distribution (KSN-P), as well as an ensemble of $10$ randomly sampled network weights (KSN-E).

\section{Discussion}

The results in Figure \ref{fig:model} demonstrate how KSNs can approximate a distribution of network weights without a 2-fold increase in the number of parameters. From the results tabulated in Section~\ref{sec:experiments}, our KSN model achieves a higher classification accuracy than the conventional BNN in all classification tasks. Most notably, in the CIFAR$10$ (Table~\ref{table:CIFAR10}) and CIFAR$100$ (Table~\ref{table:CIFAR100}) classification tasks, a $\delta$ of $0.25$ was sufficient to outperform the standard BNN. This means that the proposed KSN could achieve higher classification accuracy than BNNs with only $15\%$ of the required parameters. Moreover, KSNs consistently produce better calibrated models overall compared to BNNs, as evident by the lower NLL, ECE, ACE and MCE in Tables \ref{table:MNIST}, \ref{table:FMNIST}, \ref{table:CIFAR10}, and \ref{table:CIFAR100}.

The VOGN method could also outperform the standard BNNs with significantly fewer parameters. However, VOGN often failed to match the performance of MC-Drop and our proposed KSN method; particularly on the FMNIST (Table \ref{table:FMNIST}), CIFAR10 (Table \ref{table:CIFAR10}), and CIFAR100 (Table \ref{table:CIFAR100}) classification tasks.

Overall, MC-Drop consistently achieves high classification accuracy and can produce well calibrated models. The performance improvements with MC-Drop are most prominent on CIFAR10 (Table \ref{table:CIFAR10}), and CIFAR100 (Table \ref{table:CIFAR100}) classification tasks, where the other methods, including the proposed KSN method, suffer significant performance degradation compared to the baseline. However, since the MC-Drop method is unable to quantify uncertainty over the entire weight space, the benefits of MC-Drop for performing variational inference are limited to certain tasks (Section~\ref{sect:Introduction}).

\section{Conclusions}
The proposed KSN method can significantly reduce the number of parameters required for BDL (compared to conventional BNNs) without compromising on performance. While conventional BNNs require a 2-fold increase in the number of parameters compared to its DNN equivalent, BDL can be applied using KSNs with up to $39\%$ fewer parameters than the equivalent DNN. This enables BDL to scale up to the depths of modern DNNs.

\newpage
\vskip 0.2in
\bibliography{bib}
\end{document}